\documentclass{article}
\usepackage{iclr2026_malgai,times}

\usepackage{amsmath,amsfonts,bm}

\def\eqref#1{equation~\ref{#1}}

\def\1{\bm{1}}

\DeclareMathAlphabet{\mathsfit}{\encodingdefault}{\sfdefault}{m}{sl}
\SetMathAlphabet{\mathsfit}{bold}{\encodingdefault}{\sfdefault}{bx}{n}

\usepackage{hyperref}
\usepackage{url}
\usepackage{booktabs}
\usepackage{amsfonts}
\usepackage{nicefrac}
\usepackage{microtype}
\usepackage{graphicx}
\usepackage{xcolor}
\usepackage{algorithm}
\usepackage{algorithmic}
\usepackage{subcaption}
\usepackage{fancyvrb}
\usepackage{amsthm}
\usepackage{multirow}

\newcommand{\vmao}{\textsc{VMAO}}

\newcommand{\eg}{\textit{e.g.}}

\title{Verified Multi-Agent Orchestration: A Plan-Execute-Verify-Replan Framework for Complex Query Resolution}

\author{%
\textbf{Xing Zhang\textsuperscript{1}} \quad \textbf{Yanwei Cui\textsuperscript{1}} \quad \textbf{Guanghui Wang\textsuperscript{1}} \quad \textbf{Wei Qiu\textsuperscript{2}} \quad \textbf{Ziyuan Li\textsuperscript{2}} \\
\textbf{Fangwei Han\textsuperscript{2}} \quad \textbf{Yajing Huang\textsuperscript{2}} \quad \textbf{Hengzhi Qiu\textsuperscript{2}} \quad \textbf{Bing Zhu\textsuperscript{2}} \quad \textbf{Peiyang He\textsuperscript{1}}\thanks{Corresponding author: \texttt{peiyan@amazon.com}} \\[0.5em]
\textsuperscript{1}AWS Generative AI Innovation Center \quad \textsuperscript{2}HSBC
}

\iclrfinalcopy

\begin{document}

\maketitle

\begin{abstract}
We present \textbf{Verified Multi-Agent Orchestration (\vmao{})}, a framework that coordinates specialized LLM-based agents through a verification-driven iterative loop. Given a complex query, our system decomposes it into a directed acyclic graph (DAG) of sub-questions, executes them through domain-specific agents in parallel, verifies result completeness via LLM-based evaluation, and adaptively replans to address gaps. The key contributions are: (1) dependency-aware parallel execution over a DAG of sub-questions with automatic context propagation, (2) verification-driven adaptive replanning that uses an LLM-based verifier as an orchestration-level coordination signal, and (3) configurable stop conditions that balance answer quality against resource usage. On 25 expert-curated market research queries, \vmao{} improves answer completeness from 3.1 to 4.2 and source quality from 2.6 to 4.1 (1--5 scale) compared to a single-agent baseline, demonstrating that orchestration-level verification is an effective mechanism for multi-agent quality assurance.
\end{abstract}
\section{Introduction}

Large language models (LLMs) have enabled a new generation of multi-agent systems where specialized agents collaborate to solve complex tasks. A central challenge in such systems is \textit{coordination}: given a complex query that requires information from heterogeneous sources and diverse analytical expertise, how should agents be organized and assigned to sub-tasks? How can we ensure result quality without constant human oversight? When should the system stop iterating and synthesize a final answer? These questions are especially acute in domains like \textit{market research}, where analysts gather data from internal databases, public filings, news sources, and competitor reports, then synthesize findings into actionable insights. Information is scattered across heterogeneous sources, analysis requires diverse expertise (financial, operational, competitive), and synthesis demands cross-referencing while resolving contradictions.

Existing multi-agent frameworks fall short of these requirements. Debate-style approaches where agents critique each other's outputs \citep{du2023improving} improve reasoning quality but lack structured task decomposition. Role-playing frameworks where agents assume personas \citep{li2023camel} enable collaboration but provide no mechanism for verifying completeness. More recent systems like AutoGen \citep{wu2024autogen} and MetaGPT \citep{hong2023metagpt} offer flexible interaction patterns, yet still lack principled quality verification and adaptive refinement---critical requirements for production deployment where outputs must be reliable without constant human oversight.

We introduce \textbf{Verified Multi-Agent Orchestration (\vmao{})}, a framework that addresses these gaps through three key contributions:
\begin{enumerate}
    \item \textbf{DAG-Based Query Decomposition and Execution}: Complex queries are decomposed into sub-questions organized as a directed acyclic graph (DAG), enabling dependency-aware parallel execution with automatic context propagation from upstream results.
    \item \textbf{Verification-Driven Replanning}: An LLM-based verifier evaluates result completeness at the orchestration level, triggering adaptive replanning when gaps are identified---providing a principled coordination signal that is decoupled from individual agent implementations.
    \item \textbf{Configurable Stop Conditions}: Termination decisions are based on completeness thresholds, confidence scores, and resource constraints, enabling explicit quality-cost tradeoffs.
\end{enumerate}
On 25 expert-curated market research queries, \vmao{} improves answer completeness from 3.1 to 4.2 and source quality from 2.6 to 4.1 (1--5 scale) compared to single-agent and static multi-agent baselines.
\section{Related Work}

\textbf{Multi-Agent Coordination and Tool Use.} Recent surveys \citep{wang2024survey, xi2023rise} document the rapid growth of LLM-based multi-agent systems, which vary in coordination strategy: AutoGen \citep{wu2024autogen} uses conversational patterns, CAMEL \citep{li2023camel} employs role-playing, MetaGPT \citep{hong2023metagpt} enforces software engineering workflows, and HuggingGPT \citep{shen2023hugginggpt} orchestrates specialized models via a central controller. Orthogonally, work on tool use has focused on single-agent settings: ReAct \citep{yao2023react} established the thought-action-observation paradigm, Toolformer \citep{schick2023toolformer} enables self-supervised tool learning, and ToolLLM \citep{qin2023toolllm} scales to 16,000+ APIs. These lines of work address coordination and tool use separately, but production systems require both: multiple specialized agents, each with domain-specific tools, working in concert.

\textbf{Planning, Decomposition, and Verification.} Chain-of-Thought \citep{wei2022chain}, Tree-of-Thoughts \citep{yao2023tree}, and Least-to-Most prompting \citep{zhou2022least} decompose complex reasoning into structured steps, but operate within a single LLM rather than distributing sub-tasks across specialized agents. For output quality, Self-Consistency \citep{wang2022self} aggregates multiple reasoning paths, Self-Refine \citep{madaan2023self} iterates on single outputs, and Reflexion \citep{shinn2023reflexion} uses verbal reinforcement---all operating at the individual response level. Missing from prior work is verification at the \textit{orchestration level}: evaluating whether collective results from multiple agents adequately address the original query, and triggering targeted replanning when gaps are detected.

\textbf{Agentic Search and Deep Research.} Recent commercial systems have demonstrated the potential of multi-step agentic research: search-augmented assistants like Perplexity iteratively refine queries to synthesize information from web sources, while deep research features in frontier models \citep{openai2025deepresearch} perform extended multi-step investigation. These systems demonstrate the value of iterative research loops but are closed-source, making their coordination mechanisms difficult to study or reproduce. Our work provides an open, modular framework where the coordination strategy---particularly the verification-driven replanning loop---is explicit and configurable.

\textbf{Our Approach.} \vmao{} synthesizes these threads into a unified framework for complex query resolution. We decompose queries into a DAG of sub-questions assigned to domain-specific agents, execute them in parallel with dependency-aware scheduling, verify collective completeness via LLM-based evaluation, and adaptively replan to address gaps. We evaluate \vmao{} on market research tasks, maintaining verifiable output quality through explicit coordination mechanisms.
\section{Framework Architecture}

\subsection{Overview}

\begin{figure}[t]
\centering
\includegraphics[width=\textwidth]{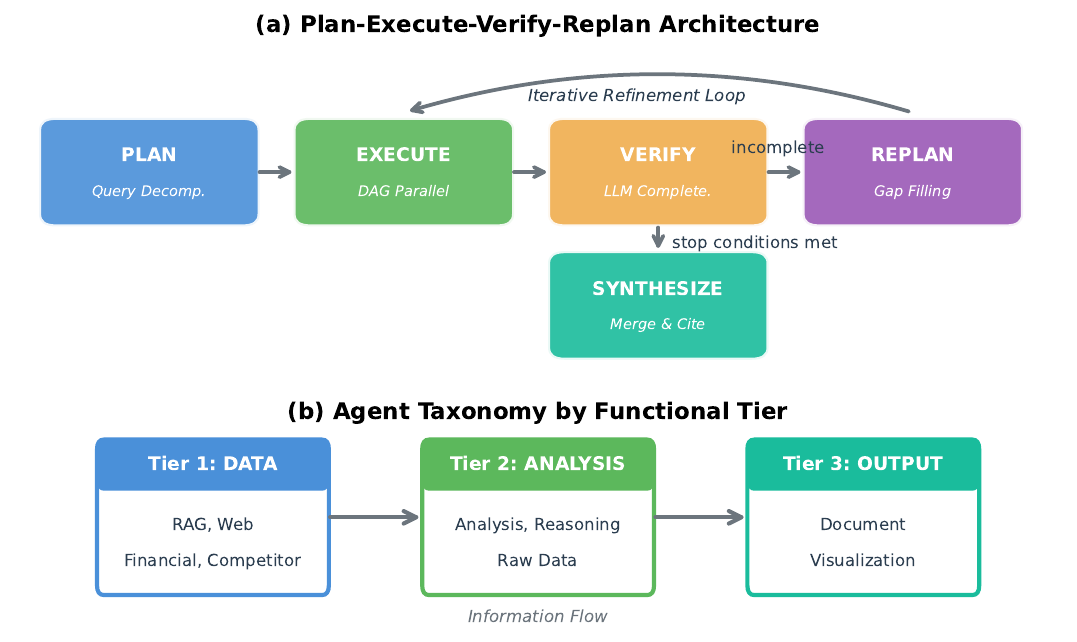}
\caption{(a) \vmao{} framework architecture showing the iterative Plan-Execute-Verify-Replan loop. (b) Agent taxonomy organized by functional tier with information flow from data gathering through analysis to output generation.}
\label{fig:framework}
\end{figure}

\vmao{} operates through five phases: \textbf{Plan}, \textbf{Execute}, \textbf{Verify}, \textbf{Replan}, and \textbf{Synthesize} (Figure~\ref{fig:framework}a). Given a complex query, the system first decomposes it into sub-questions with assigned agent types and dependencies. It then executes these through specialized agents in parallel where dependencies permit. The verify phase evaluates completeness and identifies gaps. If deficiencies exist, the system replans by generating new sub-questions or marking incomplete ones for retry. This loop continues until stop conditions are met, triggering synthesis of a final answer with proper source attribution.

Agents are organized into three functional tiers (Figure~\ref{fig:framework}b): Tier 1 (Data Gathering) agents retrieve information from diverse sources, Tier 2 (Analysis) agents reason over this data, and Tier 3 (Output) agents produce final deliverables. This hierarchy reflects the natural information flow in research tasks and enables principled task assignment by the planner.

\subsection{Planning and Execution}

The \textbf{QueryPlanner} decomposes a complex query into sub-questions organized as a DAG (Table~\ref{tab:subquestion}). An LLM identifies distinct information requirements, assigns each to an appropriate agent type, establishes dependencies where one sub-question requires another's output, and sets execution priorities.

\begin{table}[t]
\caption{Sub-question structure generated by the QueryPlanner}
\label{tab:subquestion}
\centering
\begin{tabular}{@{}p{3.5cm}p{8.5cm}@{}}
\toprule
\textbf{Field} & \textbf{Description} \\
\midrule
id & Unique identifier (e.g., sq\_001) \\
question & Specific, answerable question text \\
agent\_type & Agent from taxonomy to handle this question \\
dependencies & IDs of sub-questions that must complete first \\
priority & Execution priority (1--10); higher = more important \\
context\_from\_deps & Whether to include dependency results in prompt \\
verification\_criteria & Criteria for determining answer completeness \\
\bottomrule
\end{tabular}
\end{table}

\begin{figure}[t]
\centering
\includegraphics[width=\textwidth]{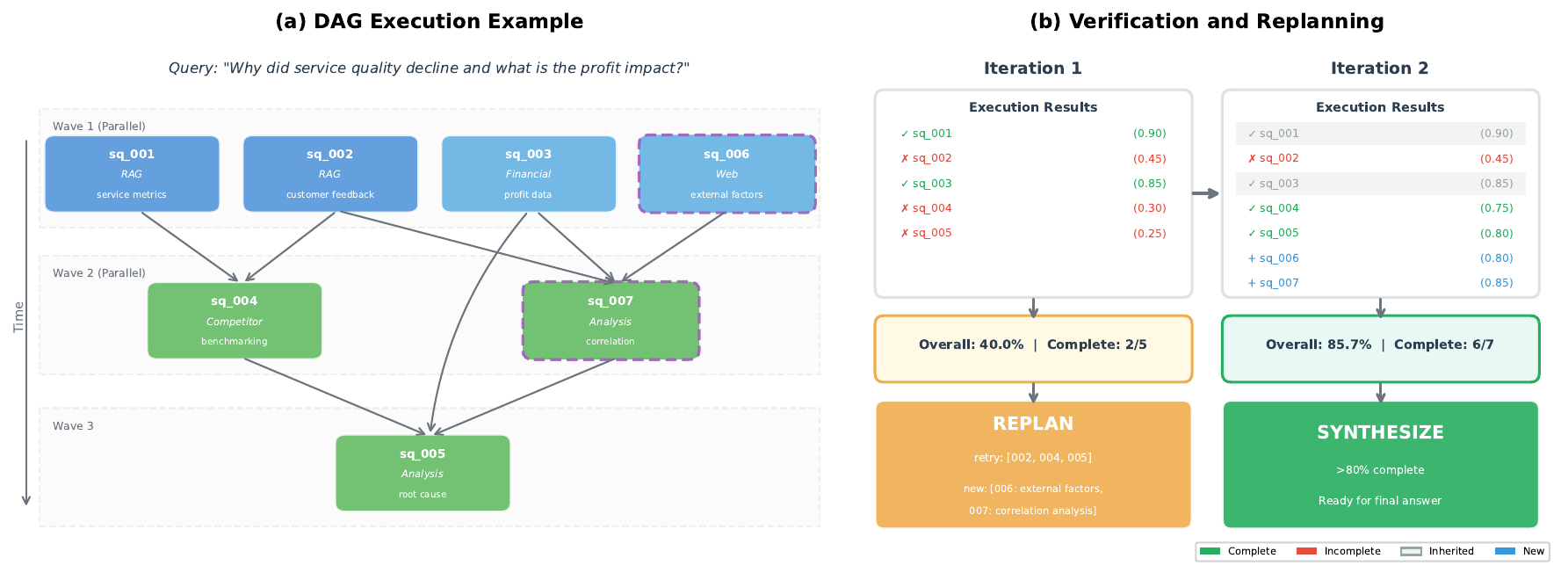}
\caption{(a) DAG execution: independent sub-questions execute in Wave 1; dependent questions in subsequent waves. (b) Verification-driven iteration: Iteration 1 identifies incomplete results, triggering replanning; Iteration 2 achieves sufficient completeness for synthesis.}
\label{fig:dag_execution}
\end{figure}

The \textbf{DAGExecutor} then orchestrates execution while respecting dependencies and maximizing parallelism (Algorithm~\ref{alg:dag}). It iteratively identifies ready questions---those whose dependencies have completed---and executes batches in parallel (default $k=3$). For sub-questions with \texttt{context\_from\_deps} enabled, results from dependencies are prepended to the query. Figure~\ref{fig:dag_execution}a illustrates how independent sub-questions execute concurrently in Wave 1, while dependent questions execute in subsequent waves. Each execution is wrapped with a configurable timeout (default: 600s) and a tool call limiter to prevent infinite loops.

\begin{algorithm}[t]
\caption{DAG-Based Parallel Execution}
\label{alg:dag}
\begin{algorithmic}[1]
\REQUIRE Execution plan $P = (Q, G)$, max concurrent $k$
\ENSURE Results $R = \{r_1, ..., r_n\}$
\STATE $completed \leftarrow \emptyset$
\WHILE{$|completed| < |Q|$}
    \STATE $ready \leftarrow \{q \in Q : deps(q) \subseteq completed \land q \notin completed\}$
    \STATE $batch \leftarrow \text{top-}k(ready, \text{by}=priority)$
    \STATE $results \leftarrow \text{parallel\_execute}(batch)$
    \FOR{$(q, r)$ in $results$}
        \IF{$q.context\_from\_deps$}
            \STATE $r \leftarrow \text{enrich\_with\_context}(r, \{R[d] : d \in deps(q)\})$
        \ENDIF
        \STATE $R[q.id] \leftarrow r$; \quad $completed \leftarrow completed \cup \{q\}$
    \ENDFOR
\ENDWHILE
\RETURN $R$
\end{algorithmic}
\end{algorithm}

\subsection{Verification, Replanning, and Synthesis}

The \textbf{ResultVerifier} evaluates whether execution results adequately answer their sub-questions (Figure~\ref{fig:dag_execution}b). For each result, it produces: status (complete/partial/incomplete), completeness score (0--1), missing aspects, contradictions, and a recommendation (accept/retry/escalate). Results already marked complete are reused to avoid redundant LLM calls.

When verification identifies gaps, the \textbf{AdaptiveReplanner} determines corrective actions: \textit{retry} sub-questions with low scores while preserving previous results, introduce \textit{new queries} to address specific missing aspects, or \textit{merge} results from multiple attempts. A key feature is result preservation---previous results are stored and merged with retry attempts, enabling progressive refinement without losing earlier findings.

Determining when to stop iterating is critical for balancing quality and cost. We introduce five configurable stop conditions (Table~\ref{tab:stop_conditions}), evaluated after each verification phase: completeness threshold (80\% of sub-questions answered), high confidence with partial coverage, diminishing returns ($<$5\% improvement), token budget (1M tokens), and maximum iterations (3). When any condition is met, the system proceeds to synthesis.

\begin{table}[t]
\caption{Stop conditions for orchestration termination}
\label{tab:stop_conditions}
\centering
\begin{tabular}{@{}llp{5.5cm}@{}}
\toprule
\textbf{Condition} & \textbf{Threshold} & \textbf{Rationale} \\
\midrule
Ready for Synthesis & 80\% complete & Sufficient sub-questions answered \\
High Confidence & 75\% conf, 50\% complete & High reliability despite partial coverage \\
Diminishing Returns & $<$5\% improvement & Further iteration yields minimal gain \\
Token Budget & 1M tokens & Hard cost limit \\
Max Iterations & 3 iterations & Hard iteration limit \\
\bottomrule
\end{tabular}
\end{table}

For large result sets ($>$15K characters or 10+ results), direct synthesis would exceed context limits. We address this through hierarchical synthesis: group results by agent type, synthesize within each group to produce condensed summaries, then integrate group summaries into a coherent final answer with proper source attribution.
\section{Implementation}

We implement \vmao{} using LangGraph for workflow orchestration and the Strands Agent framework for agent execution, integrated with AWS Bedrock. Agent execution uses Claude Sonnet 4.5 as the primary model with Claude Haiku 4.5 as a fallback for graceful degradation; verification and evaluation use Claude Opus 4.5 to provide an independent quality signal. Agents access tools through the Model Context Protocol (MCP), which exposes domain-specific capabilities via independent HTTP microservices. This modular architecture allows adding new tools without modifying agent code.

Table~\ref{tab:tools} shows the agent taxonomy with tool allocation across eight MCP servers (42 unique tools total). Each server runs independently, enabling horizontal scaling and fault isolation. Agents automatically select appropriate tools based on sub-question requirements.

\begin{table}[t]
\caption{Agent taxonomy with tool allocation across MCP servers (42 unique tools total)}
\label{tab:tools}
\centering
\begin{tabular}{@{}lllp{7.5cm}@{}}
\toprule
\textbf{Tier} & \textbf{Agent} & \textbf{Tools} & \textbf{Key Capabilities} \\
\midrule
\multirow{4}{*}{1: Data}
    & RAG & 13 & Semantic, keyword, and hybrid retrieval; metadata filtering \\
    & Web Search & 4 & General and AI-powered search, news retrieval \\
    & Financial & 7 & Stock quotes, technical indicators, fundamentals \\
    & Competitor & 11 & Market positioning, benchmarks, competitor news \\
\midrule
\multirow{3}{*}{2: Analysis}
    & Analysis & 20 & Survey analytics, financial and competitor analysis \\
    & Reasoning & 24 & Cross-domain reasoning with RAG, web, and financial tools \\
    & Raw Data & 1 & Python execution (pandas, matplotlib) \\
\midrule
\multirow{2}{*}{3: Output}
    & Document & 4 & Report generation, tables, source citations \\
    & Visualization & 6 & Chart generation, statistical summaries \\
\bottomrule
\end{tabular}
\end{table}

For production deployment, we implement several safety mechanisms: tool call limiters prevent infinite loops (max 10 consecutive same-tool calls, 50 total per agent), per-execution timeouts enforce bounded latency (default 600s), and phase-level token tracking enables budget enforcement. When the primary model (Sonnet 4.5) is unavailable, the system falls back to Haiku 4.5 with graceful degradation. Real-time observability is provided through Server-Sent Events that stream execution progress to the frontend.
\section{Experiments}

\subsection{Dataset: Market Research Queries}

We evaluate \vmao{} on \textbf{market research tasks}---a domain where traditional research typically requires 2--4 weeks of human effort. These tasks are challenging because relevant data is scattered across heterogeneous sources, answering questions requires diverse expertise (financial, operational, competitive), and synthesis demands cross-referencing while resolving contradictions. We curated 25 queries from domain experts spanning four categories:
\begin{itemize}
    \item \textbf{Performance Analysis} (8 queries): Operational metrics, trends, and causal factors. \textit{Example}: ``What factors explain the year-over-year change in customer satisfaction?''
    \item \textbf{Competitive Intelligence} (7 queries): Comparison with industry peers and market positioning. \textit{Example}: ``How does our market share compare to regional competitors?''
    \item \textbf{Financial Investigation} (5 queries): Financial metrics combined with operational context. \textit{Example}: ``What is driving the change in revenue per customer?''
    \item \textbf{Strategic Assessment} (5 queries): Open-ended synthesis across multiple dimensions. \textit{Example}: ``What are the key risks and opportunities for geographic expansion?''
\end{itemize}
Query complexity varies from simpler queries (3--5 sub-questions, 2--3 agent types) to complex ones (8--12 sub-questions, 5+ agent types with multi-level dependencies). Each query consumes 500K--1.1M tokens and requires 10--20 minutes of execution plus domain expert review, making 25 queries a practical yet meaningful evaluation set.

\subsection{Baselines and Configuration}

We compare three configurations:

\begin{itemize}
    \item \textbf{Single-Agent}: One reasoning agent with access to all tools, relying on internal reasoning to determine tool invocation order.
    \item \textbf{Static Pipeline}: Predefined agent sequence (RAG $\rightarrow$ Web $\rightarrow$ Financial $\rightarrow$ Analysis $\rightarrow$ Synthesis) without verification or replanning.
    \item \textbf{\vmao{}}: Full framework with dynamic decomposition, parallel execution, verification-driven replanning, and stop conditions.
\end{itemize}

All configurations use Claude Sonnet 4.5 for agent execution and the same tool set. We evaluate \textit{Completeness} (how thoroughly all query aspects are addressed, 1--5 scale) and \textit{Source Quality} (proper citation and traceability, 1--5 scale). Evaluation follows a two-stage process: an LLM judge (Claude Opus 4.5) first scores each response using structured rubrics, then human domain experts review and adjust scores where the LLM assessment appears inconsistent or misses domain-specific nuances. We deliberately use a different, more capable model for evaluation than for execution to reduce self-evaluation bias, though both models belong to the same family. In practice, human reviewers adjusted fewer than 15\% of LLM scores, typically by $\pm$0.5 points, indicating reasonable LLM-human alignment on these metrics.

We evaluate \textit{Completeness} rather than accuracy because deep research queries have no single ground truth---a question like ``What factors explain declining satisfaction?'' admits multiple valid answers. Completeness measures whether all relevant aspects are addressed with supporting evidence, better capturing the exploratory nature of research. Source Quality ensures answers are grounded in verifiable sources.
\subsection{Results}

Table~\ref{tab:results} presents the main results across all 25 queries. \vmao{} achieves substantially higher completeness (+35\%) and source quality (+58\%) compared to Single-Agent. The Static Pipeline improves over Single-Agent but cannot adapt when initial agents return insufficient results. \vmao{}'s verification-driven approach identifies gaps and adaptively replans, leading to more complete answers with better source attribution. The increased resource usage reflects verification overhead, justified by quality improvements.

\begin{figure}[t]
\centering
\includegraphics[width=\textwidth]{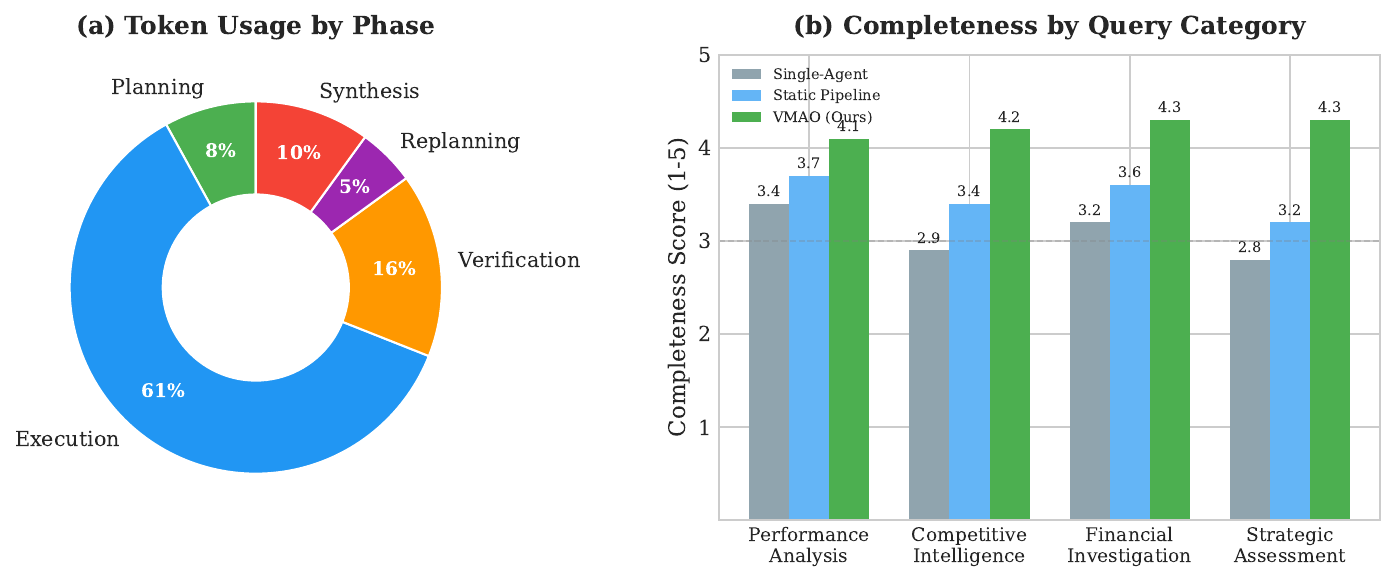}
\caption{(a) Token usage breakdown by orchestration phase for a typical query. Execution dominates at 61\%, while verification and synthesis remain efficient. (b) Completeness scores by query category across methods. \vmao{} shows consistent improvements, with largest gains on Strategic Assessment (+53\%).}
\label{fig:results}
\end{figure}

Figure~\ref{fig:results}(a) shows a typical token distribution across orchestration phases: execution dominates (61\%) as agents invoke tools and process results, while verification and synthesis remain efficient. \vmao{} demonstrates consistent improvements across all query categories (Figure~\ref{fig:results}(b)), with the largest gains on Strategic Assessment queries (+53\% completeness), which require synthesizing information across multiple dimensions. Performance Analysis queries show more modest gains, as these often have well-defined data sources that even single agents can locate.

\begin{table}[t]
\caption{Comparison of orchestration methods on market research tasks. Completeness and Source Quality are co-scored by LLM and human evaluators (1--5 scale, higher is better).}
\label{tab:results}
\centering
\begin{tabular}{@{}lcccc@{}}
\toprule
\textbf{Method} & \textbf{Completeness} & \textbf{Source Quality} & \textbf{Avg Tokens} & \textbf{Avg Time (s)} \\
\midrule
Single-Agent & 3.1 & 2.6 & 100K & 165 \\
Static Pipeline & 3.5 & 3.2 & 350K & 420 \\
\textbf{\vmao{} (Ours)} & \textbf{4.2} & \textbf{4.1} & 850K & 900 \\
\bottomrule
\end{tabular}
\end{table}

In our experiments, most queries ($>$75\%) terminate via resource-based conditions (diminishing returns, max iterations, or token budget), reflecting conservative thresholds that prioritize thoroughness over speed. These parameters are configurable for deployments requiring faster responses or lower costs.

\textbf{Evaluation Limitations.} We acknowledge three caveats: (1) 25 queries is a modest evaluation set without reported confidence intervals, (2) the LLM judge (Opus 4.5) belongs to the same model family as the execution model (Sonnet 4.5), potentially introducing shared biases despite human review, and (3) the Static Pipeline baseline tests verification and replanning jointly without a component-level ablation. We view the current evaluation as a meaningful signal of the framework's potential, while acknowledging that larger-scale evaluation with independent judges would strengthen the conclusions.
\section{Discussion}

Unlike skill-based systems (\eg, AutoGPT plugins) that invoke capabilities sequentially within a single agent, \vmao{} offers explicit DAG decomposition for interpretable plans, parallel execution reducing latency, verification-driven iteration for progressive refinement, and cross-agent synthesis with source attribution. The LLM-based verification serves as a principled coordination signal---assessing whether collective results satisfy the query---decoupling coordination from agent implementation.

\textbf{When Does Verification Help Most?} The largest gains from verification-driven replanning appear on open-ended, multi-dimensional queries (Strategic Assessment: +53\% completeness) where initial decomposition inevitably misses relevant aspects. For narrower queries with well-defined data sources (Performance Analysis), single agents already locate most relevant information, and the marginal benefit of replanning is smaller. This suggests verification is most valuable when the query space is difficult to fully characterize upfront---precisely the setting where static pipelines fail. We also observe that the majority of replanning actions are \textit{retries} of incomplete sub-questions rather than introduction of entirely new ones, indicating that agent execution variance (tool failures, insufficient search results) is a larger contributor to gaps than poor initial decomposition.

\textbf{Limitations.} Our framework has several limitations beyond the evaluation caveats noted in Section~5.3. LLM-based verification may miss subtle factual errors or hallucinations, as it evaluates completeness rather than accuracy---the verifier can confirm that a claim is present and sourced, but cannot independently establish its truth. Poor query decomposition can propagate errors downstream: if the planner misframes a sub-question, the verifier may accept a well-sourced but irrelevant answer. The system's 8.5$\times$ token cost relative to a single agent (850K vs.\ 100K tokens) may be prohibitive for latency-sensitive or cost-constrained settings. Finally, all experiments use a single model family (Claude); the framework's effectiveness with other LLM families remains untested.

\textbf{Transferability and Future Work.} The core components---DAG decomposition, verification, and replanning---are domain-agnostic and should transfer to domains like legal discovery or scientific literature review with appropriate agent and tool configuration. Future directions include learning-based stop conditions trained on execution traces, component-level ablation studies to isolate the contribution of each framework element, evaluation with diverse model families, and human-in-the-loop verification for high-stakes queries.
\section{Conclusion}

We presented \vmao{}, a framework that coordinates specialized LLM agents through a Plan-Execute-Verify-Replan loop. On 25 market research queries, \vmao{} improves answer completeness from 3.1 to 4.2 and source quality from 2.6 to 4.1 (1--5 scale) compared to single-agent baselines, with the largest gains on open-ended queries that require multi-dimensional synthesis. Our results suggest that orchestration-level verification---where an independent model evaluates whether collective agent results satisfy the original query---is an effective coordination mechanism for multi-agent systems. Key open questions remain around component-level contributions, generalization across model families and domains, and scalable evaluation methodology. We will release the implementation upon publication.

\bibliographystyle{iclr2026_conference}

\appendix
\section{Prompt Templates}
\label{app:prompts}

We provide simplified versions of the core prompts. Each follows a structured format with input specifications, decision rules, and JSON output schemas.

\vspace{0.5em}
\noindent\fbox{\parbox{0.96\columnwidth}{\small\sffamily
\textbf{Planning Prompt}\\[0.3em]
You are a query planner. Decompose complex queries into sub-questions for specialized agents.\\[0.3em]
\textit{Input:} Original query, conversation context, available agents\\[0.3em]
\textit{Planning Rules:}\\
\hspace*{1em}-- RAG First: Always search internal knowledge base first or in parallel\\
\hspace*{1em}-- Maximize Parallelism: Execute independent questions simultaneously\\
\hspace*{1em}-- Minimize Dependencies: Only when results feed into other questions\\
\hspace*{1em}-- Be Specific: Clear, answerable scope for each question\\[0.3em]
\textit{Sub-question Fields:} id, question, agent\_type, dependencies, priority, context\_from\_deps, verification\_criteria\\[0.3em]
\textit{Output:} JSON with sub\_questions array and explanation
}}

\vspace{0.8em}
\noindent\fbox{\parbox{0.96\columnwidth}{\small\sffamily
\textbf{Verification Prompt}\\[0.3em]
Verify if the sub-question has been adequately answered with proper metadata.\\[0.3em]
\textit{Input:} Sub-question, verification criteria, result, dependency results\\[0.3em]
\textit{Evaluation Criteria:}\\
\hspace*{1em}-- Completeness: All aspects of question addressed?\\
\hspace*{1em}-- Evidence Quality: Multiple sources? Cross-referenced?\\
\hspace*{1em}-- Metadata: Source attribution (filename/URL/date) present?\\
\hspace*{1em}-- Specificity: Concrete facts/numbers vs vague claims?\\
\hspace*{1em}-- Contradictions: Conflicts between sources?\\[0.3em]
\textit{Output:} JSON with verification\_status (complete/partial/incomplete), completeness\_score (0--1), missing\_aspects, confidence, recommendation (accept/retry/escalate)
}}

\vspace{0.8em}
\noindent\fbox{\parbox{0.96\columnwidth}{\small\sffamily
\textbf{Replanning Prompt}\\[0.3em]
Determine next actions based on verification results.\\[0.3em]
\textit{Input:} Original query, execution plan, completed/incomplete results, iteration count\\[0.3em]
\textit{Critical Rule:} MUST include ALL incomplete sub-question IDs in retry list.\\[0.3em]
\textit{Decision Logic:}\\
\hspace*{1em}-- completeness $>$ 0.8: Proceed to synthesis (done)\\
\hspace*{1em}-- Incomplete results exist: Add ALL to retry\_sub\_questions\\
\hspace*{1em}-- completeness 0.5--0.8: Add new\_sub\_questions to fill gaps\\
\hspace*{1em}-- Contradictions found: Add queries targeting different sources\\
\hspace*{1em}-- iterations $\geq$ max: Return empty lists (done)\\[0.3em]
\textit{Output:} JSON with retry\_sub\_questions, new\_sub\_questions, explanation
}}

\vspace{0.8em}
\noindent\fbox{\parbox{0.96\columnwidth}{\small\sffamily
\textbf{Synthesis Prompt}\\[0.3em]
Synthesize results from multiple agents into a concise, well-cited answer.\\[0.3em]
\textit{Input:} Original query, sub-question results, verification summary\\[0.3em]
\textit{Required Structure:}\\
\hspace*{1em}1. Executive Summary (2--3 sentences with key metrics)\\
\hspace*{1em}2. Key Findings (5--8 bullets with source citations)\\
\hspace*{1em}3. Analysis (2--3 paragraphs connecting insights)\\
\hspace*{1em}4. Conclusions (confidence level and limitations)\\[0.3em]
\textit{Citation Format:} [source - section/URL, metadata]\\[0.3em]
\textit{Output:} JSON with answer, key\_findings, confidence, sources, gaps
}}

\section{Configuration Parameters}
\label{app:config}

Table~\ref{tab:config} lists the default configuration parameters used in our experiments. These can be tuned for different quality-latency tradeoffs.

\begin{table}[h]
\caption{Configuration parameters for \vmao{} orchestration}
\label{tab:config}
\centering
\begin{tabular}{llp{5.5cm}}
\toprule
\textbf{Parameter} & \textbf{Default} & \textbf{Description} \\
\midrule
max\_iterations & 3 & Maximum replanning iterations \\
token\_budget & 1M & Maximum tokens before stopping \\
ready\_threshold & 0.8 & Completeness ratio for synthesis \\
high\_confidence & 0.75 & Confidence threshold for early stop \\
diminishing\_returns & 0.05 & Minimum improvement to continue \\
max\_concurrent & 3 & Parallel agent executions \\
agent\_timeout & 600s & Per-agent timeout \\
\bottomrule
\end{tabular}
\end{table}

\end{document}